\documentclass{article}

\usepackage{arxiv}

\usepackage[utf8]{inputenc} 
\usepackage[T1]{fontenc}    
\usepackage{hyperref}       
\usepackage{url}            
\usepackage{booktabs}       
\usepackage{amsfonts}       
\usepackage{nicefrac}       
\usepackage{microtype}      
\usepackage{lipsum}		
\usepackage{graphicx}
\usepackage{subcaption}
\usepackage{amsmath}
\usepackage{amssymb}
\usepackage{xcolor}
\usepackage{caption}

\newcommand{\y}{\mathbf{y_n}} 
\newcommand{\yh}{\mathbf{\hat{y}_n}}
\newcommand{\w}{\mathbf{w}}

\title{Adaptive Loss Function for Super Resolution Neural Networks Using Convex Optimization Techniques}

\author{
  Seyed Mehdi Ayyoubzadeh \\
  Department of Electrical and Computer Engineering\\
  McMaster University\\
  Hamilton, Ontario \\
  \texttt{ayyoubzs@mcmaster.ca} \\
   \And
 Xiaolin Wu \\
  Department of Electrical and Computer Engineering\\
  McMaster University\\
  Hamilton, Ontario \\
  \texttt{xwu@mcmaster.ca} \\
}

\begin{document}
\maketitle

\begin{abstract}%
	Single Image Super-Resolution (SISR) task refers to learn a mapping from low-resolution images to the corresponding high-resolution ones. This task is known to be extremely difficult since it is an ill-posed problem. Recently, Convolutional Neural Networks (CNNs) have achieved state of the art performance on SISR. However, the images produced by CNNs do not contain fine details of the images. Generative Adversarial Networks (GANs) aim to solve this issue and recover sharp details. Nevertheless, GANs are notoriously difficult to train. Besides that, they generate artifacts in the high-resolution images. In this paper, we have proposed a method in which CNNs try to align images in different spaces rather than only the pixel space. Such a space is designed using convex optimization techniques. CNNs are encouraged to learn high-frequency components of the images as well as low-frequency components.  We have shown that the proposed method can recover fine details of the images and it is stable in the training process.
	
\end{abstract}


\section{Introduction}

One of the challenging tasks in computer vision is SISR. It refers to restoring the high-resolution (HR) image from the corresponding low-resolution (LR) image. SISR is a difficult task since it is an ill-posed problem. In other words, each patch in the high-resolution image can have multiple corresponding low-resolution patches \cite{Park_2018_ECCV}. SISR has a wide range of applications, including medical imaging, aerial and satellite imaging and security and surveillance imaging \cite{10.5120/ijca2016911458}. \\
Currently, CNNs achieve state-of-the-art performance in many of the computer vision areas, such as image classification, segmentation, object detection, etc. \cite{1512.03385, Chollet2017, Huang2017, NIPS2012_4824}. CNNs can learn the highly non-linear mapping from the input space to the output space. This makes them well suited for SISR. \\
Recently, CNNs are used for SISR, and they have improved the peak-signal-to-noise-ratio (PSNR) over traditional methods \cite{1707.02921, 1808.08718, Park_2018_ECCV} by a significant margin. \cite{Dong2014, 1501.00092} proposed a simple architecture of CNN called SRCNN for SISR trained by Mean Squared Error as the loss function. Most of the CNNs use pixelwise Mean Squared Error (MSE) as the loss function to optimize the CNN. MSE can be interpreted as a Maximum Likelihood (ML) Estimator of the mean of a conditional Gaussian distribution \cite{1808.03344}. This underlying assumption about MSE is the reason for blurry high-resolution images. In fact, the CNNs trained by pixelwise MSE are not able to recover fine details and textures in the high-resolution images.
To resolve this issue, more complicated models use a combination of multiple losses to produce sharper images \cite{1808.03344, Johnson2016}. In \cite{1809.00961} Krishna et al. have proposed a loss function which is a combination of MSE loss in the pixel space and the MSE loss on the edge space. They have used Canny operator to derive the edge map of a high-resolution image. Although this loss function can perform better than the plain MSE, yet it can not recover different frequencies and textures of the high-resolution images. \\ 
\cite{1603.08155} proposed a new loss function called perceptual loss as the auxiliary loss function for training SISR CNNs. Since perceptual loss considers the MSE between the features extracted by VGGNet \cite{1409.1556}, the CNN trained by this loss can recover more textures in the high-resolution images.
Some people have used the loss function of Generative Adversarial Networks (GANs) (GAN loss) \cite{1406.2661} for SISR task \cite{1609.04802} as the auxiliary loss function. Although using GANs is a great idea to produce crisp images, they are notoriously tricky and unstable to train \cite{1801.04406}. Also, they can introduce unpleasant artifacts in the high-resolution images. Moreover, for complex discriminator architectures, the training time can be prohibitively long since the only way to train the network is the first-order optimization methods.   
In this paper, we proposed a method in which we use the core idea of GAN i.e., looking for a space in which the discrimination of the output of the CNN and the ground truth is maximized, but instead of having a CNN as the discriminator, a set of filter is designed using promising convex optimization techniques. The proposed approach has several advantages:
\begin{itemize}
	\item {It can handle explicit constraints (prior knowledge) for designing the filters that can emphasize on certain textures or high-frequency details}
	\item {Using convex optimization techniques in designing filters make the method fast and efficient}
	\item {It can recover fine textures and edges in the high-resolution images}
	\item {The training process is stable}
\end{itemize}

\section{Proposed Method}
In this section, we describe the proposed method to update the loss function of super-resolution CNN. First, the common loss function for CNN is analyzed, then the proposed method is presented for updating the loss function of CNN.

\subsection{Loss function of the Super Resolution CNN}
The typical loss function of the super-resolution CNNs is the pixelwise mean squared error. Let $\mathbf{y_n}$ and $\mathbf{\hat{y}_n}$ denote the high-resolution ground truth image and the output of the CNN for the $n$th image in the dataset respectively. The mean squared error can be written as:
\begin{equation}
L(\w)=\frac{1}{N}\sum_{n=1}^{N}\|\y-\yh(\w)\|^2 
\end{equation}
Where $\mathbf{w}$ represents all the weights of the CNN and $N$ is the total number of the images in the dataset. the pixelwise MSE can lead to blurry outputs of the CNN. To overcome this issue, we create a new space and measure MSE. We assume that this space is created by convolving the images with different filters of a filter bank $\mathcal{F}$ which consists of $M$ filters ($\mathcal{F}=\{F_1, F_2, \cdots, F_M\}$). The resulted loss function of the CNN is:
\begin{equation}
L(\w)=\frac{1}{N}\sum_{n=1}^{N}\|\y-\yh(\w)\|^2 + \frac{\alpha}{MN}\sum_{n=1}^{N}\sum_{m=1}^{M}\|F_m*\y-F_m*\yh(\w)\|^2
\label{eqn:cnnloss}
\end{equation}
Where $F_m$ is the $m$th filter of the filter bank out of $M$ filters and $\alpha$ is the coefficient of the auxiliary loss function. The CNN minimizes \ref{eqn:cnnloss} concerning its parameters via backpropagation.

\subsection{Designing Discriminator Filter Bank ($\mathcal{F}$)}
The purpose of the discriminator filter bank is to transform the images and the outputs of the CNN such that the difference between these two would be maximized, i.e.,

\begin{equation}
\operatorname*{maximize}_{F_m} \sum_{m=1}^{M}\sum_{n=1}^{N}\|F_m*(\y-\yh(\w))\|^2
\label{eqn:maxterm}
\end{equation}

To reduce the computational burden of this optimization problem we design the filters based on $N_s$ samples of the dataset, i.e.,
\begin{equation}
\operatorname*{maximize}_{F_m} \sum_{m=1}^{M}\sum_{n=1}^{N_s}\|F_m*(\y-\yh(\w))\|^2
, \;\; F_m\in \mathbb{R}^{k\times k}, \y, \yh \in \mathbb{R}^{w\times h} 
\label{eqn:maxterm2}
\end{equation}
Where $k$ is the size of the filters, $w$ and $h$ are the output images dimensions.
To simplify \ref{eqn:maxterm2}, we define $Y_n$ as $\y-\yh$. Therefore, we have:
\begin{equation}
\operatorname*{maximize}_{F_m} \sum_{m=1}^{M}\sum_{n=1}^{N_s}\|F_m*Y_n\|^2
, \;\; F_m\in \mathbb{R}^{k\times k}, Y_n \in \mathbb{R}^{w\times h} 
\label{eqn:maxterm3}
\end{equation}
For further simplification of \ref{eqn:maxterm3}, we write the convolution as the matrix multiplication. Let $D_{Y_n}$ and $\tilde{F}_m$ denote the Doubly block circulant matrix of $Y_n$ and the vectorized filter $F_m$. 

\begin{equation}
\operatorname*{maximize}_{\tilde{F}_m} \sum_{m=1}^{M}\sum_{n=1}^{N_s}\|D_{Y_n}\tilde{F}_m\|^2
,\widetilde{F}_m\in \mathbb{R}^{k^2},D_{Y_n}\in \mathbb{R}^{l\times k^2},l=(w+k-1)\times (h+k-1)
\label{eqn:maxterm4}
\end{equation}

Therefore, we can write \ref{eqn:maxterm4} in the quadratic form as follows:
\begin{equation}
\operatorname*{maximize}_{\tilde{F}_m} \sum_{m=1}^{M}\sum_{n=1}^{N_s} \tilde{F}_m^T D_{Y_n}^T D_{Y_n}\tilde{F}_m
\label{eqn:maxterm5}
\end{equation}

\subsubsection{Imposing Constraints on the Filters}
To design a meaningful filter bank, we need to impose some constraints on the problem \ref{eqn:maxterm5}. In the following, we explain the constraints on the filters. Scaling the elements of a filter by a constant factor does not change the behavior of the filter. Therefore, we need to have a unit-energy filter that does to not change the amplitude of the signal. In other words, the desired filters have the unit norm:
\begin{equation}
\|\widetilde{F}_m\|_2^2=1 \;\; \forall m
\end{equation}
Secondly, we need various filters in the filter bank, and each filter tries to extract certain textures or frequencies from the outputs. To achieve this goal, we assume that the filters in the filter bank are mutually near orthogonal, i.e, 
\begin{equation}
|\widetilde{F}_i^T\widetilde{F}_j| \leq \epsilon \;\; \forall i,j, i\neq j
\end{equation}
Lastly, the filters in the filter bank are expected to be bandpass or high pass filters. In other words, the filters should not be low pass filters. We can write this constraint as follows:
\begin{equation}
\sum_{a=1}^{k^2}\widetilde{F}_{m}^{(a)}=0 \;\; \forall m
\label{eqn:mainProblem}
\end{equation}
Where $\widetilde{F}_m^{(a)}$ is the $a$th element of $\widetilde{F}_m$. 

\subsubsection{Iterative Design of the Filters}
To handle the orthogonality constraint, we design filters one by one and then add orthogonality constraint. In this case, the problem will be a nonconvex quadratically constrained quadratic (QCQP). To develop $m$th filter, we use the method described in \cite{Luo2010} to convert inhomogeneous QCQP
to the homogeneous one:

\begin{equation}
\begin{aligned}
& \underset{\tilde{F}_m}{\text{minimize}}
& & -\sum_{n=1}^{N_s} \widetilde{F}_m^TD_{Y_n}^TD_{Y_n}\widetilde{F}_m \\
& \text{subject to}
& &\begin{matrix}\begin{pmatrix}\widetilde{F}_m^T & t_m\end{pmatrix}\\\mbox{}\end{matrix}\begin{pmatrix} \widetilde{F}_m \\ t_m \end{pmatrix}=2 \;\; \forall m \\
& & & t_m^2=1 \\
& & &  \begin{matrix}\begin{pmatrix}\widetilde{F}_m^T & t_m\end{pmatrix}\\\mbox{}\end{matrix}
\begin{pmatrix} 0 & \frac{\boldsymbol{1}}{2} \\ \frac{\boldsymbol{1}}{2} & 0 \end{pmatrix} 
\begin{pmatrix} \widetilde{F}_m \\ t_m \end{pmatrix}=0 \;\; \forall m \\
& & &  \begin{matrix}\begin{pmatrix}\widetilde{F}_m^T & t_m\end{pmatrix}\\\mbox{}\end{matrix}
\begin{pmatrix} 0 & \frac{\widetilde{F}_i}{2} \\ \frac{\widetilde{F}_i^T}{2} & 0 \end{pmatrix} 
\begin{pmatrix} \widetilde{F}_m \\ t_m \end{pmatrix} \leq \epsilon \;\; ,i < m \\
& & &  \begin{matrix}\begin{pmatrix}\widetilde{F}_m^T & t_m\end{pmatrix}\\\mbox{}\end{matrix}
\begin{pmatrix} 0 & \frac{\widetilde{F}_i}{2} \\ \frac{\widetilde{F}_i^T}{2} & 0 \end{pmatrix} 
\begin{pmatrix} \widetilde{F}_m \\ t_m \end{pmatrix} \geq -\epsilon \;\; ,i < m \\
\end{aligned}
\label{eqn:iterativeProblem}
\end{equation}

Now, let $C_n = -D_n^TD_n$, $x_m = \begin{pmatrix} \widetilde{F}_m \\ t_m \end{pmatrix}$ and $X_m=x_mx_m^T$, we can rewrite the problem as follows:

\begin{equation}
\begin{aligned}
& \underset{X_m}{\text{minimize}}
& &  \sum_{n=1}^{N_s}Tr(C_nX_m)  \\
& \text{subject to}
& & Tr(X_m)=2 \\
& & & Rank(X_m)=1 \\
& & & X_m \succeq 0 \\
& & &  	Tr(\begin{pmatrix} \mathbf{0} & \frac{\boldsymbol{1}}{2} \\ \frac{\boldsymbol{1}}{2} & 0 \end{pmatrix} 
X_m)=0 \;\; \\
& & &  	Tr(\begin{pmatrix} \mathbf{0} & \frac{\widetilde{F}_i}{2} \\ \frac{\widetilde{F}_i^T}{2} & 0 \end{pmatrix} 
X_m) \leq \epsilon \;\; , i<m\\
& & &  	Tr(\begin{pmatrix} \mathbf{0} & \frac{\widetilde{F}_i}{2} \\ \frac{\widetilde{F}_i^T}{2} & 0 \end{pmatrix} 
X_m) \geq -\epsilon \;\; , i<m
\end{aligned}
\label{eqn:fproblem}
\end{equation}

Where $Tr(A)$ represents the trace of matrix $A$. This problem can be solved efficiently using convex optimization techniques such as Semi Definite Relaxation (SDR) \cite{Luo2010} or Convex Concave Programming \cite{1604.02639}.

\begin{figure}
	\begin{center}		
		\includegraphics[width=0.5\linewidth]{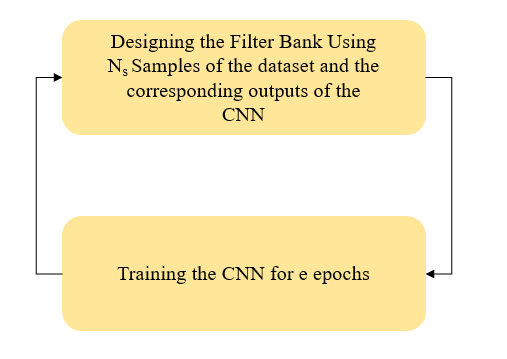}
		\caption{Overview of the proposed algorithm}
		\label{fig:alg}
	\end{center}
\end{figure}

\subsection{Overview of the Algorithm}
The block diagram of the algorithm is shown in Figure \ref{fig:alg}. First, we randomly pick $N_s$ high-resolution images from the dataset and also their corresponding prediction of the CNN. Then we solve problem \ref{eqn:fproblem} to design the discriminator filter bank. Afterward, CNN is trained with the loss function described in \ref{eqn:cnnloss} for $e$ epochs. We repeat this procedure until overfitting. 

\begin{figure}
	\includegraphics[width=\linewidth]{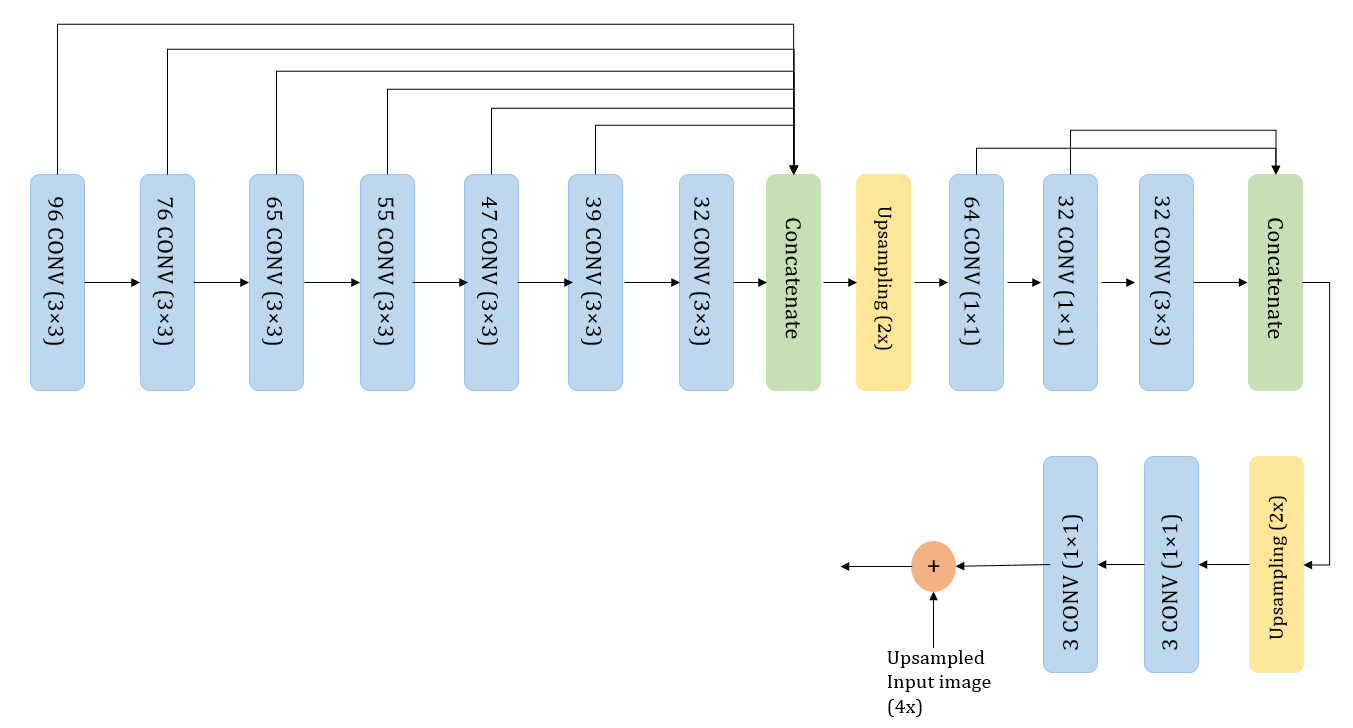}
	\caption{CNN architecture}
	\label{fig:arch1}
\end{figure}

\section{Experimental Results}
\section{Setup}
To evaluate the performance of the algorithm, the architecture is shown in Figure \ref{fig:arch1} is used. For training, the CNN DIV2K dataset \cite{Agustsson_2017_CVPR_Workshops, Timofte_2017_CVPR_Workshops} is used. The filter size of the filter bank ($k$) is $7$ and the total number of filters in the filter bank ($M$) is $32$. We have used $300$ samples from the dataset to design the discriminator filter bank ($N_s=300$). The CNN is trained for $100$ epochs with the batch size $32$ and the learning rate $10^{-3}$. The discriminator filter bank is updated after every $5$ epochs ($e=5$). We have used three metrics to evaluate the performance of the classifier: PSNR, SSIM \cite{Wang2004} and PSNR on the high pass part of the image. The first two metrics are quite common for measuring the quality of the high-resolution images. However, we need a metric to show the difference in the high-frequency components of the images. To do so, the Laplacian filter is first applied to the images and then the PSNR will be calculated. We can write:

\begin{equation}
PSNR_{High\;Frequencies}=PSNR(\nabla^2 \y, \nabla^2 \yh) 
\end{equation}
Where $PSNR(ref, img)$ is the PSNR of the output of the super-resolution network concerning the ground truth.

\subsection{Discussions}

\begin{figure}
	\centering
	\begin{subfigure}[b]{0.48\linewidth}
		\includegraphics[width=\linewidth]{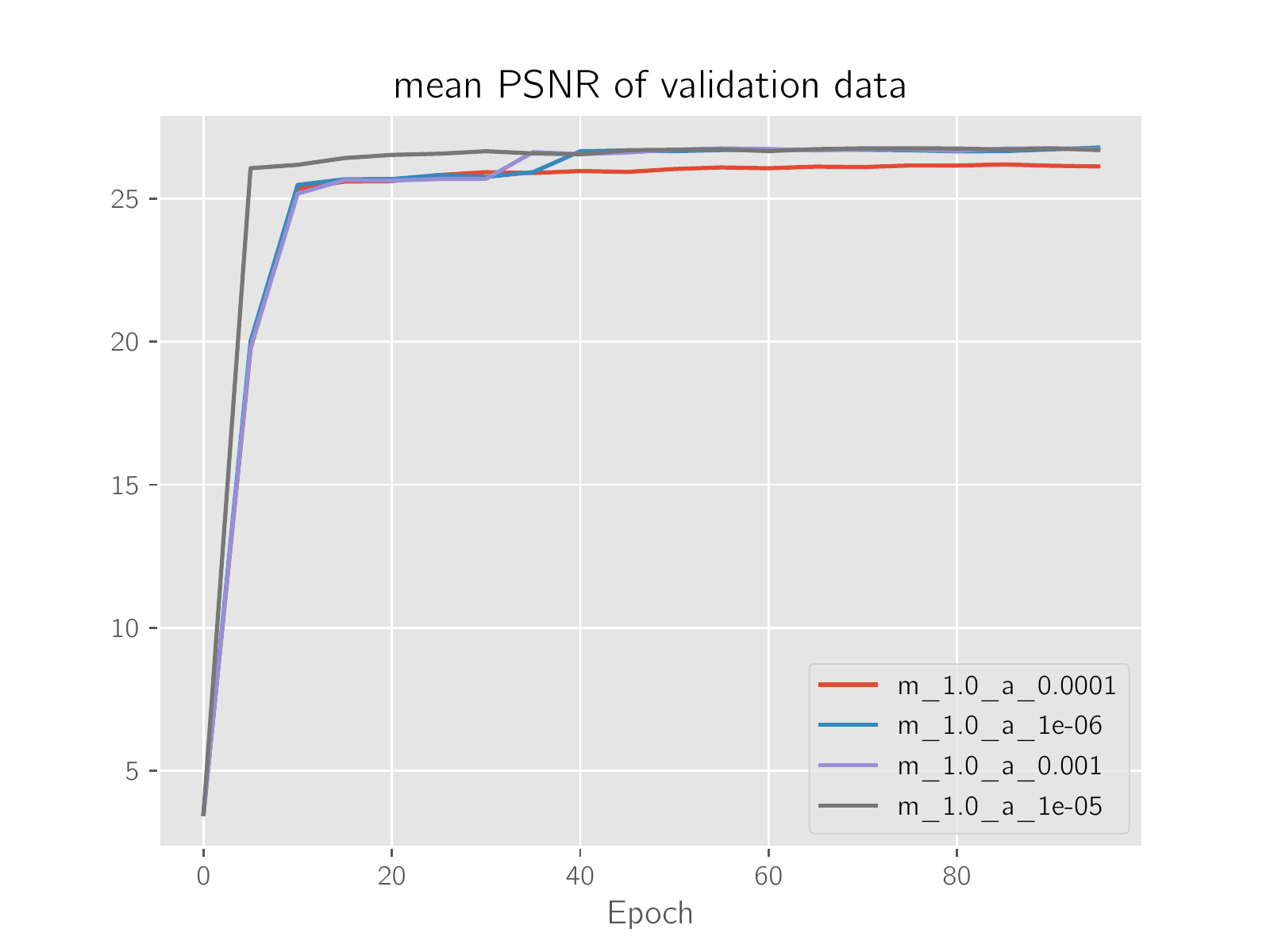}
	\end{subfigure}
	\begin{subfigure}[b]{0.48\linewidth}
		\includegraphics[width=\linewidth]{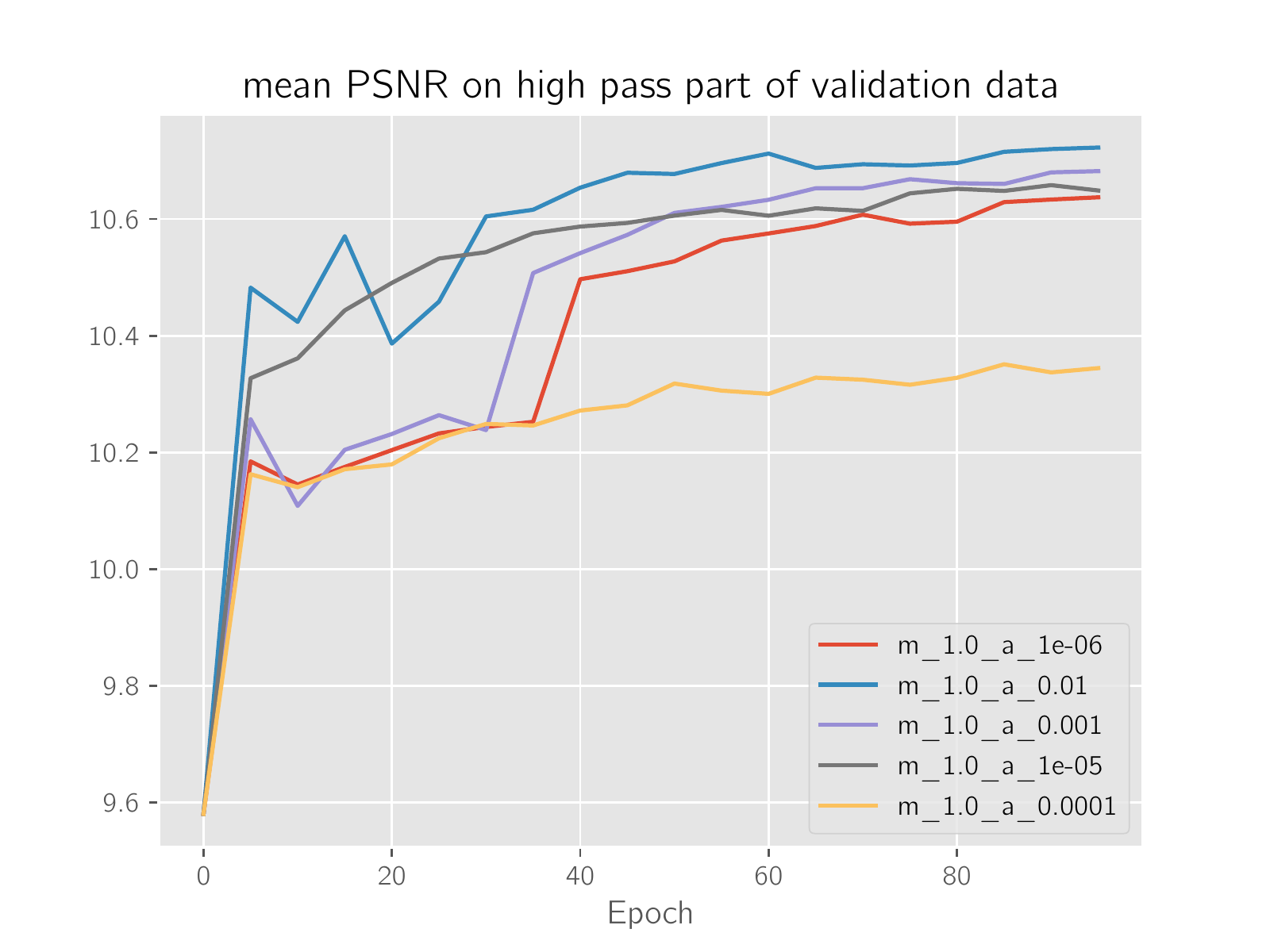}
	\end{subfigure}
	\caption{(a) Mean PSNR (b) Mean PSNR for the high frequencies}
	\label{fig:valplot}
\end{figure}

\begin{figure}
	\includegraphics[width=\linewidth]{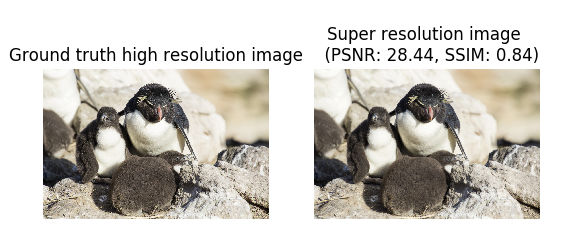}
	\caption{An example of the CNN output and the corresponding ground truth}
	\label{fig:ex1}
\end{figure}

The mean PSNR of the validation images and the mean PSNR on the high frequencies for different $\alpha$ are shown in Figure \ref{fig:valplot} (a) and (b). As shown, mean PSNR for different $\alpha$ is more or less the same, however, the CNN trained with auxiliary loss function has better PSNR for the high frequencies. In other words, the auxiliary loss function helps CNN to recover fine details and textures more accurately. An example of the CNN output is shown in Figure \ref{fig:ex1}.

\begin{figure}[!h]
	\centering
	\begin{subfigure}[b]{0.32\linewidth}
		\includegraphics[width=\linewidth]{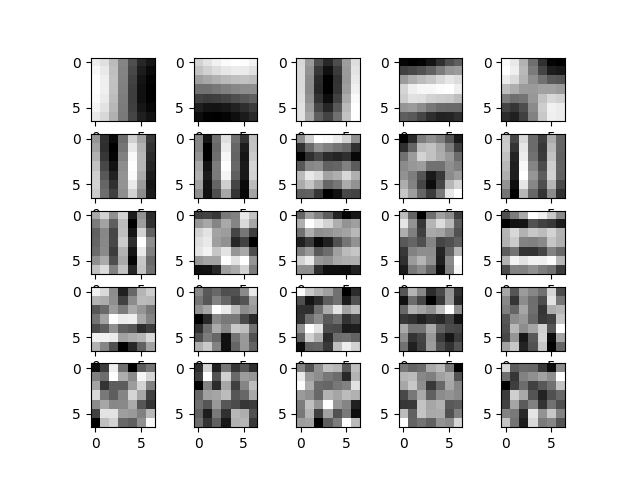}
	\end{subfigure}
	\begin{subfigure}[b]{0.32\linewidth}
		\includegraphics[width=\linewidth]{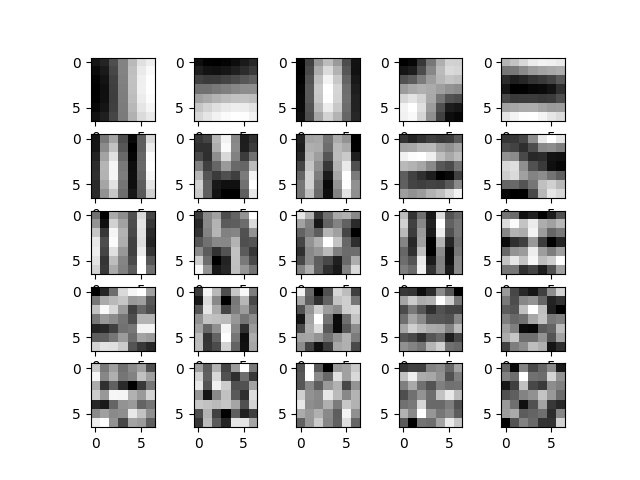}
	\end{subfigure}
	\begin{subfigure}[b]{0.32\linewidth}
	\includegraphics[width=\linewidth]{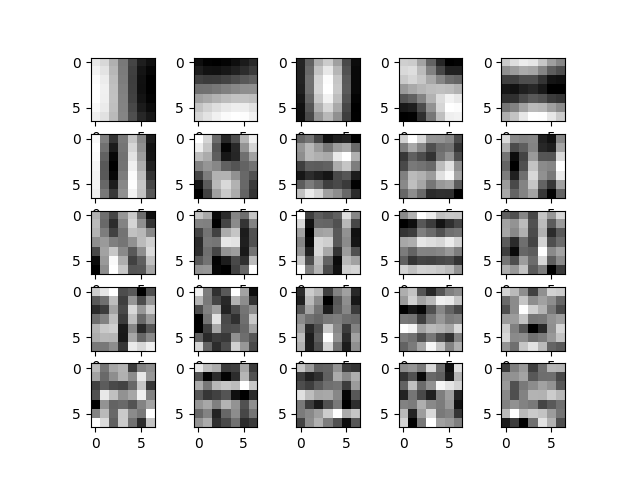}
\end{subfigure}
	\caption{Discriminating filters for the first epoch for different color channels}
	\label{fig:filte0}
\end{figure}

\begin{figure}[!h]
	\centering
	\begin{subfigure}[b]{0.32\linewidth}
		\includegraphics[width=\linewidth]{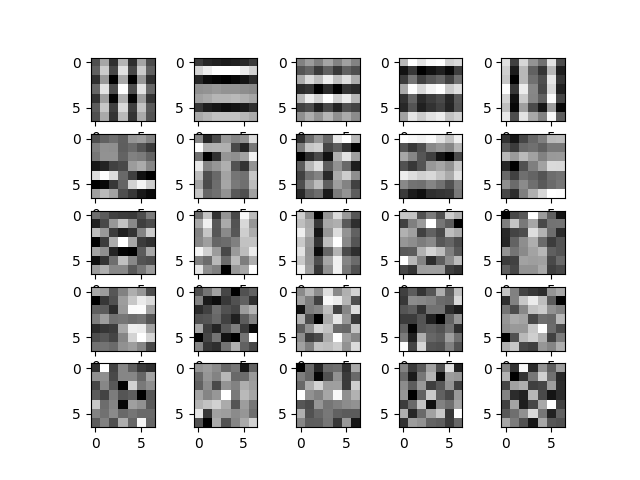}
	\end{subfigure}
	\begin{subfigure}[b]{0.32\linewidth}
		\includegraphics[width=\linewidth]{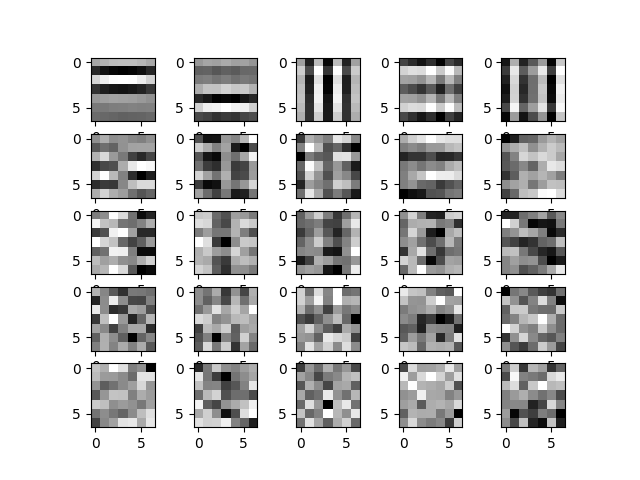}
	\end{subfigure}
	\begin{subfigure}[b]{0.32\linewidth}
		\includegraphics[width=\linewidth]{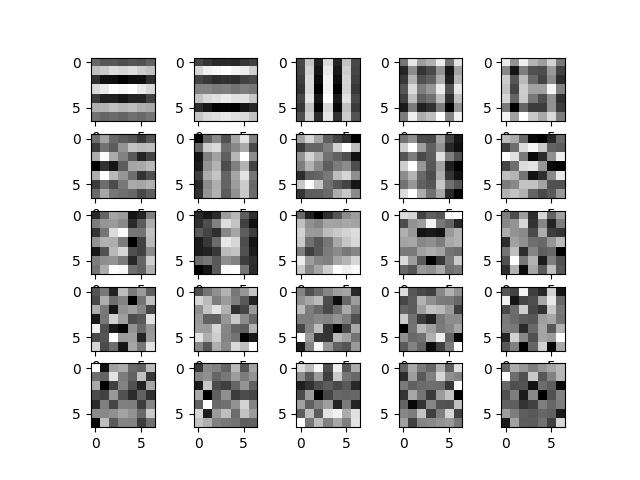}
	\end{subfigure}
	\caption{Discriminating filters for epoch $95$ for different color channels}
	\label{fig:filte95}
\end{figure}

It is beneficial to see the discriminating filters for different epochs to see their behavior. For the first and $95$th epochs, the filters for different channels are shown in Figures \ref{fig:filte0} and \ref{fig:filte95} respectively. As one can see these filters are mostly like Gabor filters useful for texture extraction in different orientations and frequencies.

\subsection{Conclusion}
In this paper, we present a new algorithm to design adaptive loss function for the super-resolution CNNs. The proposed method can encourage CNN to learn fine details and textures of the images by designing a discriminator filter bank. Designing the discriminator filter bank is done by convex optimization techniques. Using these techniques makes the optimization efficient and let us impose explicit constraints for designing discriminator filters.

\bibliographystyle{unsrt}  

\clearpage 

\bibliography{references}

\end{document}